\documentclass[runningheads]{llncs}
\usepackage[T1]{fontenc}
\usepackage{wrapfig}

\usepackage{graphicx}
\usepackage[T1]{fontenc}

\usepackage{hyperref}
\usepackage{booktabs}
\usepackage{subcaption} 
\usepackage{float}      

\usepackage{amsmath}
\usepackage{amssymb}
\usepackage{amsfonts}
\usepackage{color}

\begin{document}

\title{TransZero: Parallel Tree Expansion in MuZero using Transformer Networks}

\author{Emil Malmsten \and Wendelin Böhmer}

\institute{Delft University of Technology, The Netherlands \\
\email{emil.malmsten@outlook.com}}
\maketitle              

\begin{abstract}
We present \textbf{TransZero}\footnote{\url{https://github.com/emalmsten/TransZero}}, a model-based reinforcement learning algorithm that removes the sequential bottleneck in Monte Carlo Tree Search (MCTS). Unlike MuZero, which constructs its search tree step by step using a recurrent dynamics model, TransZero  employs a transformer-based network to generate multiple latent future states simultaneously. Combined with the Mean-Variance Constrained (MVC) evaluator that eliminates dependence on inherently sequential visitation counts, our approach enables the parallel expansion of entire subtrees during planning. Experiments in MiniGrid and LunarLander show that TransZero achieves up to an eleven-fold speedup in wall-clock time compared to MuZero while maintaining sample efficiency. These results demonstrate that parallel tree construction can substantially accelerate model-based reinforcement learning, bringing real-time decision-making in complex environments closer to practice.
\end{abstract}

\section{\label{sec:intro}Introduction}
Deep reinforcement learning has achieved major breakthroughs in sequential decision-making tasks, most notably in games. A landmark was AlphaGo \cite{alphago}, which combined deep neural networks with Monte Carlo Tree Search (MCTS) \cite{MCTS} to surpass human experts in Go. Its successor, AlphaZero \cite{alphazero}, generalized the approach to multiple board games through self-play. MuZero \cite{muzero} extended this line of work by matching or surpassing human performance not only in board games but also in Atari, while learning an implicit model of the environment without direct access to its dynamics.

Despite these advances, MuZero is limited by its sequential planning strategy: action sequences must be unrolled step by step with a recurrent dynamics network, and each expansion depends on updated visitation counts. This makes tree construction inherently sequential and restricts scalability.

We propose TransZero, to our knowledge, the first algorithm to construct MCTS trees without sequential dependencies. The TransZero algorithm was originally introduced in the author’s master’s thesis \cite{transzero}; here we provide a concise presentation of the method and results. 

Our approach has two main components: (i) a transformer-based \cite{attention_is_all} dynamics network that generates entire rollouts in parallel via self-attention, and (ii) a Mean-Variance Constrained (MVC) evaluator \cite{jaldevik} that enables node expansion independent of visitation counts. Together with minor modifications to MCTS and a custom token masking, these elements allow entire subtrees to be expanded simultaneously. This design removes the fundamental sequential bottleneck of MuZero and opens the door to scalable, parallel planning and faster training. Our experiments show that TransZero is an order of magnitude faster while maintaining sample efficiency, taking a step toward making model-based reinforcement learning practical for real-time decision-making.
 
\section{Background}\label{sec:background}

\subsection{Markov Decision Processes}\label{sec:RL}
Reinforcement learning (RL) models decision making as a Markov decision process (MDP) \cite{rl-intro}. At time $t$, the agent in state $s_t$ takes action $a_t$, receives reward $r_t$, and transitions to $s_{t+1}$. The aim is to learn a policy $\pi(s,a)$ maximizing expected return over time $V_\pi(s)=\mathbb{E}_\pi[\sum_{t=0}^\infty \gamma^t r_t \mid s_0=s]$ with discount factor $\gamma\in[0,1]$. In partially observable MDPs (POMDPs), the agent only observes $o_t$ instead of $s_t$. Model-based RL (MBRL) improves efficiency by learning dynamics and planning in a latent space.

\subsection{MuZero \label{sec:muzero}}
MuZero \cite{muzero} is a model-based reinforcement learning algorithm that integrates a learned environment model with MCTS in latent space. At each timestep $t$, MCTS is used to build a decision tree where the edges correspond to actions $a$ and the nodes to latent states $\tilde{s}$. The resulting tree guides action selection in the real environment.  

\begin{figure}
    \centering
    \includegraphics[width=1.05\textwidth]{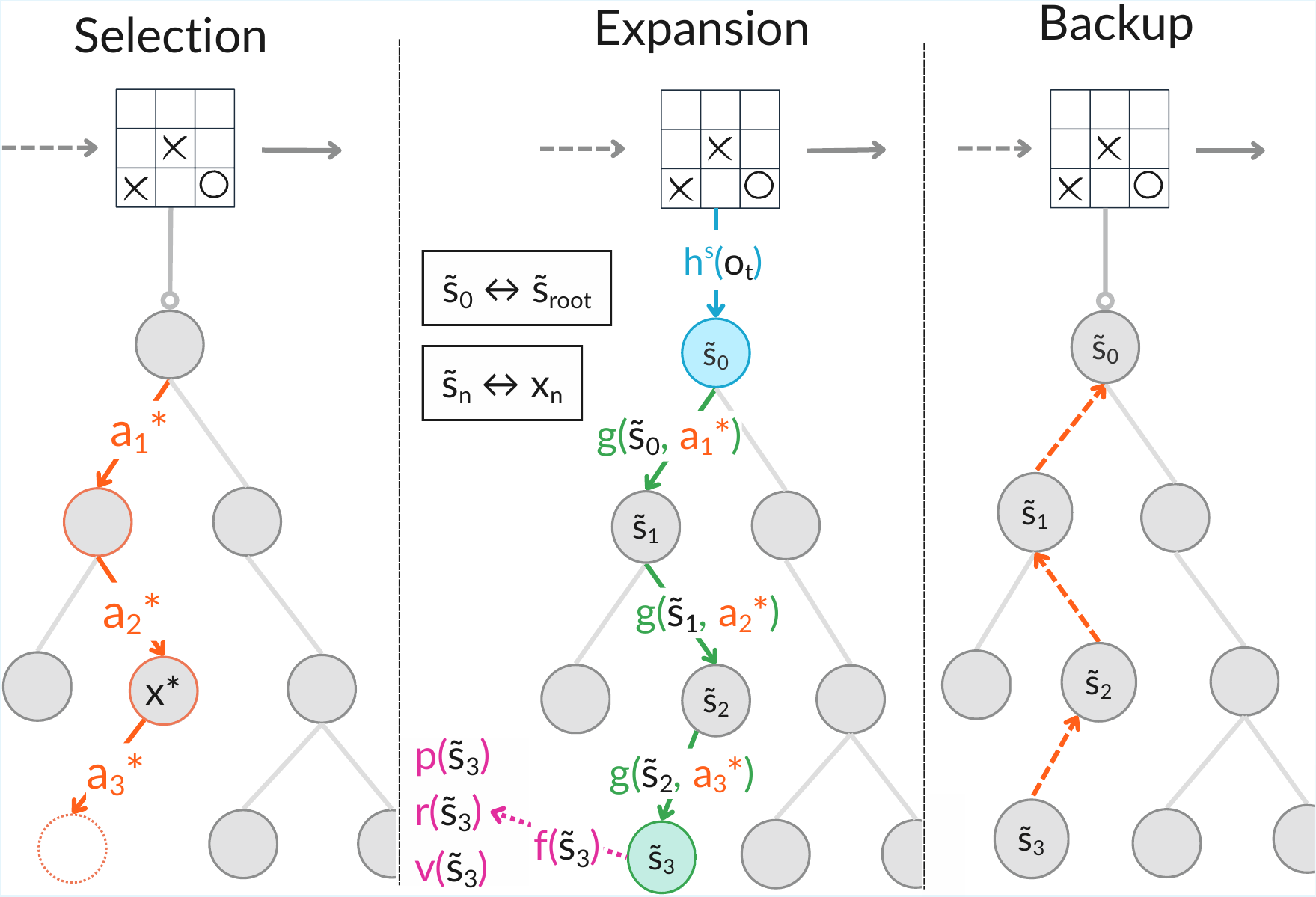}
    \caption{MuZero’s MCTS process in latent space, showing selection using PUCT, expansion through the learned dynamics, and value backup to ancestor nodes.} 
    \label{fig:muzero_mcts}
\end{figure}

Given the current observation $o_t$, the representation network produces a latent root state $\tilde{s}_0 = h^s_\theta(o_t)$. From this root, a fixed number of \emph{simulations} are run, each consisting of three phases, as illustrated in \autoref{fig:muzero_mcts}. 

(i) \emph{Selection.} Actions are chosen recursively using the PUCT rule until a leaf node is reached:  
\begin{equation}
\label{eq:puct}
a^* = \operatorname*{arg\,max}_a \Big[ Q(x \uplus a) + C_{puct} \cdot p(x,a) \frac{\sqrt{\sum_b N(x \uplus b)}}{1+N(x \uplus a)} \Big],
\end{equation}
where $x \uplus a$ is the node you reach by taking action $a$ at node $x$, $Q(x \uplus a)$ is the Q-value estimate, $N(x \uplus a)$ the visit count, $p(x,a)$ the prior policy predicted by the network, and $C_\text{puct}$ a variable balancing exploration and exploitation. We will use  $x$ and $\tilde s$ interchangeably when describing the decision tree. 

(ii) \emph{Expansion.} From the selected leaf node $\tilde{s}_n$, the dynamics network generates the next latent state $\tilde{s}_{n+1} = g_\theta(\tilde{s}_n, a)$. The prediction network then outputs value, reward, and policy estimates:  
\[
v(x),\; r(x),\; p(x) = f_\theta(x).
\]  

(iii) \emph{Backup.} The predicted value and reward are backpropagated to update the statistics of ancestor nodes. Their visit counts are also incremented, which affects the PUCT score and this is what makes the process sequential. Running full rollouts to terminal states is infeasible in complex environments, therefore the value estimate $\hat{V}$ is used as an approximation of long-term returns. These estimates are combined with observed rewards to update the action-value statistics $Q(x \uplus a)$, which in turn guide future simulations toward more promising branches.

After all simulations, MuZero selects an action in the real environment based on the visitation counts at the root. The policy is defined as  
\[
\pi(s_t, a) = \frac{N(s_t, a)^{1/\tau}}{\sum_{b \in \mathcal{A}} N(s_t, b)^{1/\tau}},
\]
where $\tau$ is a temperature parameter that controls exploration.  

The networks are then trained by unrolling $K$ steps in latent space and minimizing a combined loss over value, reward, and policy targets, derived from real trajectories and the corresponding MCTS statistics.

\subsection{General Tree Evaluation \cite{jaldevik}}\label{sec:general_mcts}
In MuZero, node evaluation is tightly linked to visitation counts, which restricts modifications to the search procedure. For example, breadth-first construction would render visit counts meaningless. To decouple evaluation from tree construction, Jaldevik \cite{jaldevik} proposed a general framework in which nodes are assessed by a \emph{tree evaluation policy} $\tilde{\pi}$. This defines the value estimate of a node as
\[
\hat{V}_{\tilde{\pi}}(x) = \sum_{a \in \mathcal{A}_v} \tilde{\pi}(x,a)\, Q_{\tilde{\pi}}(x \uplus a),
\]
with $Q_{\tilde{\pi}}(x \uplus a_v) = v(x)$ and $\mathcal{A}_v = \mathcal{A} \cup \{a_v\}$ being an extended action set, where the special simulation action $a_v$ queries the network value estimate directly.

The goal of the MVC-evaluator is to balance the maximum estimated value and the minimum variance. This is done using a parameter $\beta > 0$  as follows: 
\begin{equation}
 \label{eq:mvc}
\tilde{\pi}_{\mathrm{MVC}}(x,a) \propto \tilde{\pi}_{\mathrm{Var}}(x,a)\,\exp\big(\beta Q_{\tilde{\pi}}(x \uplus a)\big)
\end{equation}
\[
\tilde{\pi}_{\text{Var}} = \mathbb{V}\bigl[Q_{\tilde{\pi}}(x \uplus a)\bigr]^{-1}.
\]

As $\beta \to 0$, it converges to $\tilde{\pi}_{\mathrm{Var}}$; as $\beta \to \infty$, it recovers $\tilde{\pi}_Q$ which is the maximum value policy. This formulation enables node evaluation independently of visitation counts, making it compatible with alternative tree expansion strategies.

\section{The TransZero Algorithm}

We introduce \textbf{TransZero}, a MuZero variant that integrates (i) a transformer-based dynamics network, (ii) the MVC evaluator, and (iii) parallel subtree expansion in MCTS. Together, these components enable parallelized planning. 

\subsection{Transformer Dynamics}
In MuZero, the dynamics network $g_\theta$ unrolls states recurrently. We replace it with a transformer-based variant, $g_\theta^{\text{trans}}$, which generates latent states $\tilde{S}$ from the root latent state $\tilde{s}_{\text{root}}$ and a sequence of embedded actions $X^{\text{emb}}$. Each action in $X$ is mapped through a learnable embedding layer and augmented with sinusoidal positional encoding, a function we denote by $h^a_\theta$. We treat $\tilde{s}_{\text{root}}$ as the first token and $X^{\text{emb}}$ as the subsequent tokens. The transformer then outputs a sequence of latent states:
\[\tilde S = (\tilde{s}_0, \dots, \tilde{s}_n) = g_\theta^{\text{trans}}([\tilde{s}_{\text{root}} \ || \ X^{emb} ]).\]
Here, the actions ($a_{1}, \dots, a_{n}$) have been turned into latent states $(\tilde{s}_1, \dots, \tilde{s}_n)$.
 These latent states together with $\tilde s_{\text{root}}$ can then be passed to the prediction network to output value, policy prior, and reward estimates. 

Firstly, this dynamics network enables us to train over the entire unroll sequence in parallel. In the conventional setup, each new latent state is computed sequentially, with each step depending on the previous one. This requires $K$ sequential forward passes to produce the full sequence of latent states. In contrast, TransZero processes the entire action sequence in a single forward pass, producing all latent states $(\tilde{s}_0, \dots, \tilde{s}_k)$ simultaneously for the full unroll. Here, a causal mask $M$ ensures that each position in the sequence only attends to past and current tokens, preventing access to future information. 

Secondly, with some adaptations, this allows us to expand entire subtrees in parallel, as illustrated in \autoref{fig:pll_mcts}. We start by selecting the subtree to expand by selecting a node $x^*$ using PUCT as before and then using that as the root of the new subtree. The amount of subtree layers to expand $N_l$ is kept as a tunable variable. 
$X$ is set to be the concatenation of all actions leading to $x^*$ and all actions corresponding to each node in the subtree rooted in $x^*$. This is embedded as before with each action getting the positional encoding corresponding to its depth in the search tree. 

We use the mask $M_{tree} \in \{0, 1\}^{|X| \times |X|}$ to enforce that an action can only attend its ancestors, not to other unrelated nodes at the same or shallower depth. For example, in \autoref{fig:pll_mcts}, if we consider the action $a_{4,2}$, it should only be able to attend to $a_{3,1}$, $a_{2}^*$, $a_{1}^*$, and $\tilde{s}_{\text{root}}$. It must \emph{not} attend to sibling actions such as $a_{4,1}$. Formally
\[
    M_{\text{tree }ij} =
    \begin{cases}
        1 & \text{if } x_j \in \mathbf{a}_{\tilde{s}_{\text{root}} \rightarrow x_i}, \\
        0 & \text{otherwise.}
    \end{cases},
\]
where $\mathbf{a}_{\tilde{s}_{\text{root}} \rightarrow x}$ is the set of actions in the search tree leading from the root to node $x$.

The root latent state and the new $X^{emb}$ are then passed to the transformer dynamics network. This will result in the latent states for the whole subtree which is then passed as a batch to the prediction network.

Subtrees are stored as flat lists, allowing efficient indexing and parallel backup. Since Q-value and variance updates depend only on child nodes, backups across the same depth are independent and can be computed in parallel. This reduces backup complexity from exponential in branching factor to linear in subtree depth given parallel processing.

\begin{figure}
    \centering
    \includegraphics[width=0.967\textwidth]{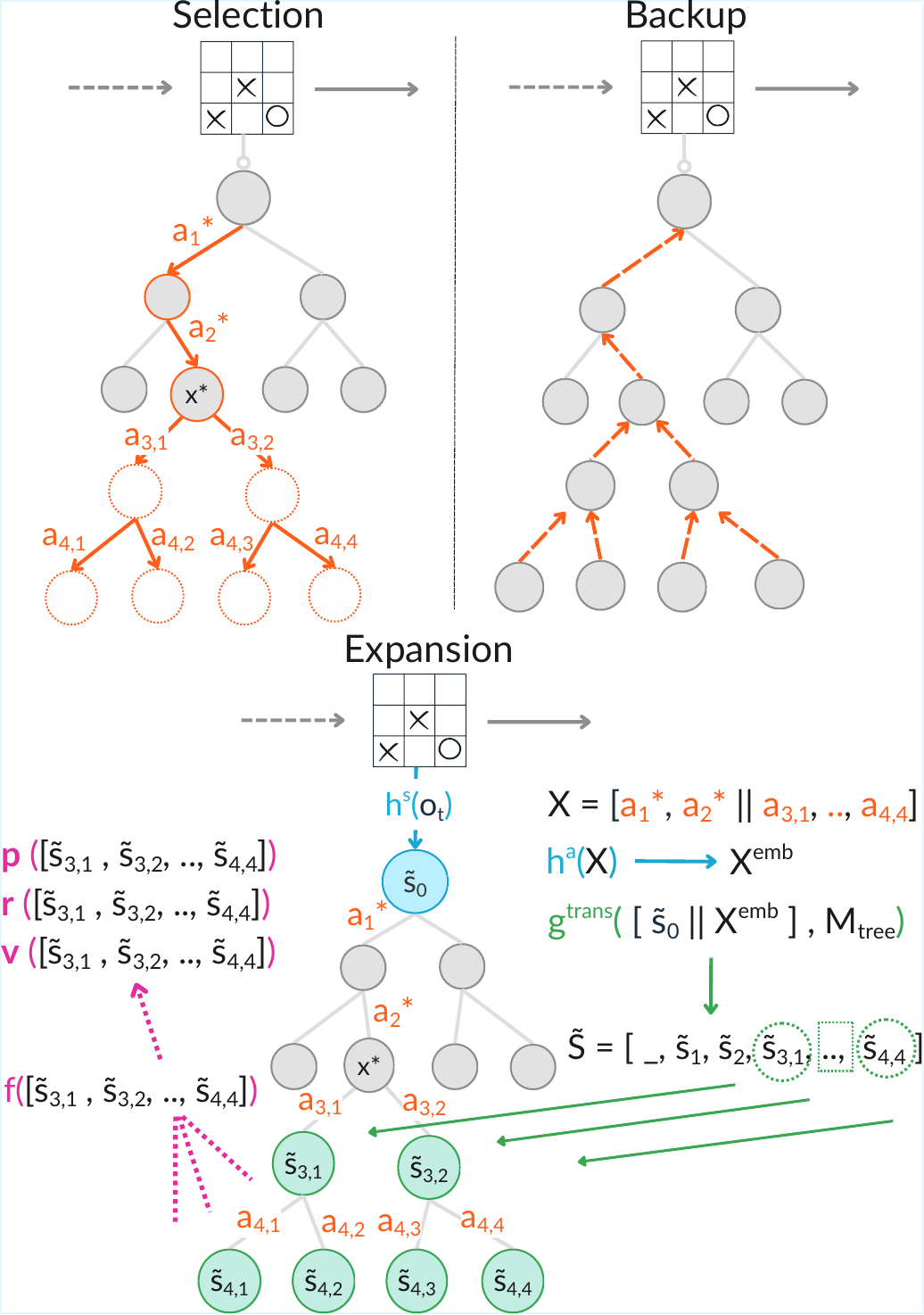}
    \caption{One simulation of MCTS in TransZero. The expansion step covers the full subtree under $x^*$. The transformer dynamics $g^{\text{trans}}_\theta$ (green) generates all latent states from $M_{tree}$ and the ordered action sequence $X$ (orange) concatenated to $\tilde{s}_0$, while the prediction network $f_\theta$ (pink) outputs rewards, values, and policy priors in batch.}
    \label{fig:pll_mcts}
\end{figure}

\subsection{MVC-Based Evaluation}
The Q-value estimate $Q_{\tilde{\pi}}$ is defined recursively as

$$
Q_{\tilde \pi}(x) = r(x) + \gamma \sum_{a \in \mathcal{A}_v} \tilde{\pi}(x, a) \, Q_{\tilde \pi}(x \uplus a),
$$

where $\tilde{\pi}(x,a) = 0$ for all $a \in \mathcal{A}$ if $\text{leaf}(x)$.
For non-leaf nodes, $\tilde{\pi}(x,a)$ is computed as described in \autoref{eq:mvc}.

The variance of the Q-value is given by
$$
\mathbb{V}[Q_{\tilde{\pi}}(x)] = \mathbb{V}[r(x)] + \gamma^2 \cdot \left(\tilde{\pi}(x, \mathcal{A}_v)^\top \tilde{\pi}(x, \mathcal{A}_v)\right) \, \mathbb{V}[Q_{\tilde{\pi}}(x \uplus \mathcal{A}_v)].
$$

For leaf nodes, the variance is defined as
$$
\mathbb{V}[Q_{\tilde{\pi}}(x)] = \mathbb{V}[r(x)] + \gamma^2 \cdot \mathbb{V}[v(x)] \quad \text{if } \text{leaf}(x).
$$

In a deterministic environment, the reward variance is zero, while the value prediction variance is set to one.

Following \autoref{eq:puct}, the PUCT score is defined as the sum of $Q(x \uplus a)$ and an exploration bonus $U(x \uplus a)$. Since MVC already defines a Q-value estimate through $Q_{\tilde{\pi}}(x \uplus a)$, we replace $Q(x \uplus a)$ accordingly. The exploration term $U(x \uplus a)$ depends on visitation counts, which we replace with $\mathbb{V}[Q_{\tilde{\pi}}(x)]$. This substitution is justified by the result that the variance of the average return is inversely proportional to the visitation count \cite{jaldevik}. The resulting formulation is

$$
U_{\tilde{\pi}}(x \uplus a) = C_{puct} \cdot p(x, a) \cdot \frac{\sqrt{\mathbb{V}[Q_{\tilde{\pi}}(x)]^{-1}}}{1 + \mathbb{V}[Q_{\tilde{\pi}}(x \uplus a)]^{-1}}.
$$

Finally, policy targets are derived from the MVC tree policy $\tilde{\pi}_{\text{MVC}}$, ensuring consistency between planning and learning.

\section{Related Work}

\subsection{Transformers in Model-Based Reinforcement Learning}  
Transformers have been increasingly applied to MBRL. TransDreamer \cite{transdreamer} extends Dreamer \cite{dreamer} by replacing its recurrent world model with a transformer state-space model, enabling better handling of long-term dependencies and improving performance on navigation tasks. IRIS \cite{iris} adopts a discrete autoencoder and an autoregressive transformer to learn environment dynamics in latent space, achieving strong performance on Atari without explicit planning. These works demonstrate the effectiveness of transformers as world models for capturing long-range temporal structure.  

\subsection{Transformers in AlphaZero and MuZero}  
Some studies have integrated transformers into AlphaZero and MuZero. Chessformer \cite{chessformer} replaces AlphaZero’s CNN-based representation network with a transformer encoder, using relative position encodings tailored to chessboard inputs and achieving superior playing strength at lower computational cost. UniZero \cite{unizero} introduces a transformer backbone in MuZero, aggregating entire sequences of past states and actions to form context-rich latent histories. By decoupling memory from state representation, UniZero improves scalability and data efficiency in both discrete and continuous control tasks.  

\subsection{Parallelizing MCTS}  
Prior work on parallel MCTS explored \emph{leaf}, \emph{root}, and \emph{tree} parallelization strategies \cite{pll_mcts}. Root parallelization, which merges multiple independent trees at the root, achieves the highest speedups with minimal coordination overhead. Tree parallelization shares a single tree across threads but requires costly synchronization, while leaf parallelization yields the smallest gains. These methods differ from our approach and are not mutually exclusive. Instead of running independent searches or synchronizing shared trees, TransZero parallelizes expansion within a single search by leveraging transformer dynamics and MVC evaluation.

\section{Experimental Evaluation}
\label{cha:results}

\subsection{Environments}
We evaluate TransZero in two domains. Firstly, \texttt{LunarLander-v3} for which we use the standard Gym implementation. Secondly, MiniGrid where we construct a custom environment. Each episode includes three randomly placed lava tiles, and both the agent’s initial position and orientation are randomized. The goal location is fixed in the lower-right corner. Successfully reaching the goal yields a reward between 5 and 10, scaled by the optimality of the agent’s trajectory. For LunarLander we run 5 random seeds and for MiniGrid 10. The full set of hyperparameters is available in the project’s GitHub repository.

\subsection{Results}
\setlength{\columnsep}{24pt} 

\begin{wraptable}{r}{0.5\columnwidth}
\vspace{-2.5em} 

\centering

\captionsetup{font=footnotesize,skip=2pt}
\caption{Final performance and relative runtime of MuZero and TransZero with standard error.}
\footnotesize
\setlength{\tabcolsep}{2pt} 

\renewcommand{\arraystretch}{0.8} 
\begin{tabular}{l|cc}
\toprule
 & \textbf{MuZero} & \textbf{TransZero} \\
\midrule
\multicolumn{3}{l}{\textbf{LunarLander}} \\
Final Reward   & 220 ($\pm$ 30)  & 220 ($\pm$ 30) \\
Runtime        & \, 1.0 ($\pm$ 0.01) & 0.092 ($\pm$ 0.003) \\
\midrule
\multicolumn{3}{l}{\textbf{MiniGrid}} \\
Final Reward   & 8.5 ($\pm$ 0.8) & 8.4 ($\pm$ 0.9) \\
Runtime        & \, 1.0 ($\pm$ 0.003) & 0.41 ($\pm$ 0.01) \\
\bottomrule
\end{tabular}

\label{tab:res}
\vspace{-0.8em} 
\end{wraptable}

The learning trends can be seen in \autoref{fig:plots} and the summary of the performance in \autoref{tab:res}. TransZero achieves sample efficiency comparable to MuZero across all tested environments, while requiring significantly less wall-clock training time. On LunarLander, TransZero completes training approximately 11$\times$ faster than MuZero, and on MiniGrid, it is 2.5$\times$ faster. 
\begin{wraptable}{r}{0.4\columnwidth} 
\vspace{-4em} 
\captionsetup{font=footnotesize,skip=2pt}
\caption{Theoretical TransZero speedup as a function of the number of MuZero (MZ) simulations}

\centering
\footnotesize 
\setlength{\tabcolsep}{4pt} 
\renewcommand{\arraystretch}{0.7} 
\begin{tabular}{cc}
\toprule
\textbf{MZ. Sims.} & \textbf{Speedup} \\
\midrule
4    & 3.0   \\
20   & 11    \\
340  & 150   \\
1640 & 560   \\
6170 & 270   \\
\bottomrule
\end{tabular}

\label{tab:1sim_table}
\vspace{-4em} 

\end{wraptable}

\subsection{Simulation Cost Analysis}
The performance gap between the two environments arises from differences in the number of simulations executed per action. In MiniGrid, TransZero uses 4 simulations (with the number of subtree layers $N_l = 2$) compared to MuZero’s 25. In LunarLander it uses 2 simulations (with $N_l = 3$) versus 50. This translates to a reduction of 6.25$\times$ simulations in MiniGrid and 25$\times$ in LunarLander compared to MuZero. Although each simulation in LunarLander expands more nodes on average (84 vs.\ 12 in MiniGrid), these expansions are processed in parallel, resulting in negligible per-simulation overhead. As shown in \autoref{tab:1sim_table}, the theoretical scaling limits of such parallel expansion become apparent: up to approximately 1640 simulations can be executed efficiently, yielding speedups of up to 560$\times$. Beyond this point, relative gains diminish. These results are measured on an RTX~4090 GPU; larger GPUs may sustain even higher levels of parallelism. 

This table shows that the reduced training time closely matches the reduced planning time, indicating that parallelization is the main cause. It also further highlights that greater advantages may be realized in environments where more than 50 simulations are needed for effective planning.

\begin{figure}[H]
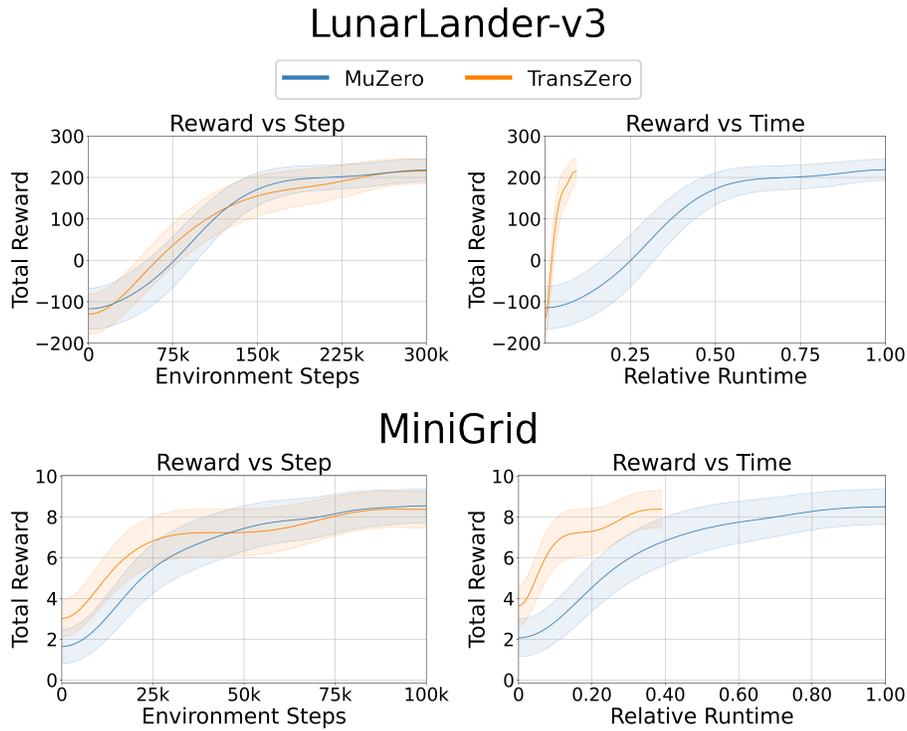


    \centering
    \vspace{0.25em} 
    \input{plots/env_steps}
    \vspace{-4.25em} 
    \input{plots/time}
    \vspace{-1.5em} 
    \setcounter{figure}{3} 

    \caption{
        Average reward of the agents on MiniGrid and LunarLander, as a function of steps (left) and as a function of relative wall-clock time (right). The shaded area shows standard error. 
    }
    \label{fig:plots}

\end{figure}

\clearpage
\section{\label{cha:conclusion}Conclusion and Future Work}
We introduced TransZero, a MuZero variant that removes the sequential bottleneck of MCTS. By replacing the recurrent dynamics model with a transformer and incorporating an MVC evaluator, TransZero enables subtree expansion in parallel rather than recurrently.

Experiments on MiniGrid and LunarLander show that TransZero maintains sample efficiency while reducing planning time by up to 11$\times$, yielding similar improvements in overall training time. These results demonstrate the scalability benefits of parallel tree construction and suggest that larger gains may be achievable. Beyond games, parallel tree construction could help make RL practical in robotics, online decision-making, and other real-world applications.

Future work includes extending TransZero to more challenging environments, as LunarLander remains relatively simple. In addition, many computations in the key–query attention matrix during subtree expansion are redundant because tokens that represent the same action at the same tree depth share identical embeddings and positional encodings. A similar approach as suggested in Perceiver \cite{perciever} with cross-attention could reduce the number of calculations exponentially. Techniques from EfficientZero \cite{efficientzero} may also improve sample efficiency.

\bibliographystyle{splncs04}
\bibliography{sources}

\clearpage

\appendix
\section{Ablations and Additional Plots}
For completeness, we also report additional experiments and ablations. We ran experiments in a MiniGrid $3 \times 3$ environment with two lava tiles and a $5 \times 5$ environment with four lava tiles. We further tested two ablated variants: TransZero-Seq, which is MuZero with the dynamics network replaced by a transformer (still using visitation counts and expanding one node at a time), and TransZero-Seq-MVC, which is the same but uses MVC for evaluation instead of visitation counts, without expanding multiple nodes in parallel. Finally, we evaluated MuZero-FC, where the ResNet is replaced by a fully connected network.

The corresponding results are shown in \autoref{fig:abl_plots}, and the relative training speeds compared to MuZero are reported in \autoref{tab:abl_speed_table}.

\begin{table}[H]
\centering
\setlength{\tabcolsep}{12pt}
\caption{Comparison of the time to complete training in MiniGrid and LunarLander. The times are relative to the times it took for MuZero.}

\begin{tabular}{l|cc}
\toprule
\textbf{Model} & MiniGrid & LunarLander  \\
\midrule
MuZero                  & 1.0 (±0.01) & 1.0 (±0.01) \\
MuZero-FC               & 0.57 (±0.01) & 0.60 (±0.01) \\
TransZero-Seq               & 0.82 (±0.01) & 0.76 (±0.01) \\
TransZero-Seq-MVC      & 1.3 (±0.02) & 1.6 (±0.08) \\
TransZero     & 0.41 (±0.00) & 0.092 (±0.00) \\ 
\bottomrule
\end{tabular}
\vspace{1em}

\label{tab:abl_speed_table}
\end{table}

\begin{figure}
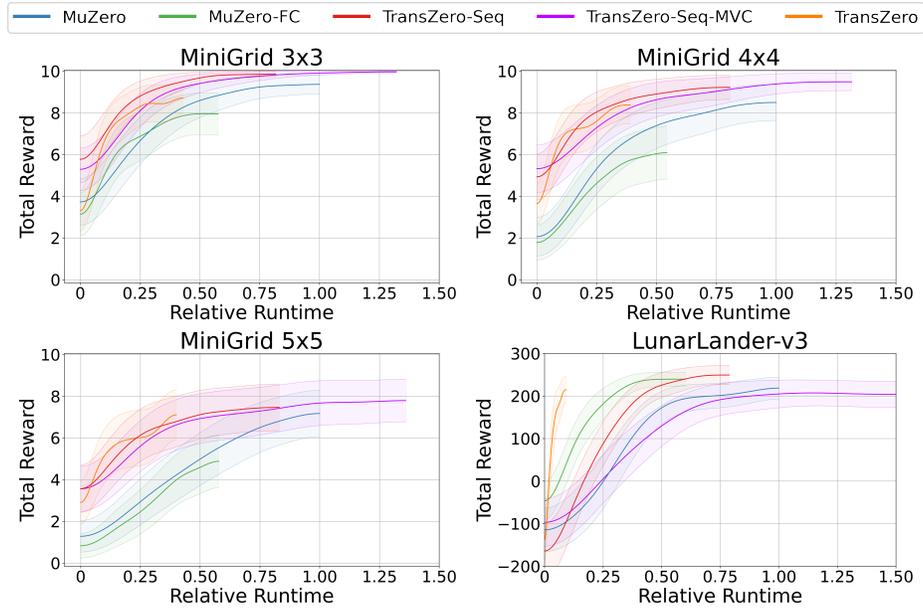

    \centering
    \input{abl_plots/env_steps}
    \vspace{-3.85em} 
    \input{abl_plots/time}
    \vspace{-3em} 
    \setcounter{figure}{4} 

    \caption{
        Average reward of the agents on LunarLander and MiniGrid environments as a function of environment steps (top) and relative wall-clock time to MuZero (bottom). The shaded area shows standard error. 
    }
    \label{fig:abl_plots}
\end{figure}

\end{document}